\documentclass[10pt,halfline,a4paper]{ouparticle}

\usepackage{authblk}
\usepackage{amsmath,amssymb,amsfonts}
\usepackage{graphicx}
\usepackage[export]{adjustbox}
\usepackage{float}
\usepackage{algorithmic}
\usepackage[bookmarks=false]{hyperref}
\usepackage{natbib}
\usepackage{booktabs}
\usepackage{soul}
\usepackage{multicol}
\usepackage{color}
\usepackage{xcolor}
\usepackage{tikz}
\usetikzlibrary{positioning}
\usepackage{tabularx}
\usepackage{graphicx}

\begin{document}
\title{Improved detection of discarded fish species through BoxAL active learning}

\author[1]{Maria Sokolova}
\author[1]{Pieter M. Blok}
\author[2]{Angelo Mencarelli}
\author[2]{Arjan Vroegop}
\author[3]{Aloysius van Helmond}
\author[1]{Gert Kootstra}
\affil[1]{Wageningen University and Research, Agricultural Biosystems Engineering group, Wageningen, 6700 AA, the Netherlands}
\affil[2]{Wageningen University and Research, Greenhouse Horticulture Unit, Wageningen, 6700 AP, The Netherlands}
\affil[3]{Wageningen University and Research, Wageningen Marine Research, IJmuiden, 1970 AB, the Netherlands}

\abstract{In recent years, powerful data-driven deep-learning techniques have been developed and applied for automated catch registration. However, these methods are dependent on the labelled data, which is time-consuming, labour-intensive, expensive to collect and need expert knowledge.
In this study, we present an active learning technique, named BoxAL, which includes estimation of epistemic certainty of the Faster R-CNN object-detection model. The method allows selecting the most uncertain training images from an unlabeled pool, which are then used to train the object-detection model. To evaluate the method, we used an open-source image dataset obtained with a dedicated image-acquisition system developed for commercial trawlers targeting demersal species. We demonstrated, that our approach allows reaching the same object-detection performance as with the random sampling using $400$ fewer labelled images. Besides, mean AP score was significantly higher at the last training iteration with $1100$ training images, specifically, $39.0\pm{1.6}$ and $34.8\pm{1.8}$  for certainty-based sampling and random sampling, respectively. Additionally, we showed that epistemic certainty is a suitable method to sample images that the current iteration of the model cannot deal with yet. Our study additionally showed that the sampled new data is more valuable for training than the remaining unlabeled data. Our software is available on \url{https://github.com/pieterblok/boxal}.  
} 

\date{\today}

\keywords{Remote electronic monitoring, sustainable fisheries, convolutional neural networks, training optimization}

\maketitle

\section{Introduction}

In the European Union (EU), the Landing Obligation has been implemented to document and land complete catch, including unmarketable catch in addition to the target catch. This regulation was introduced in 2013 and has been fully implemented in the member states in 2019 \citep{EU2015} The regulation aims at increasing fisheries transparency and documenting the amount of fish that has been caught, including the unwanted bycatch \citep{van2020electronic}. However, as the landing obligation comes with additional costs for the fishers, methods are being investigated to document the bycatch without the need for landing, to move from a landing obligation to a documentation obligation.

To estimate the amount and composition of discards onboard, the conventional method is to have human observers joining fishing trips to sample the discarded fish. These samples are then analysed manually by the experts \citep{poos2013estimating}, making it time consuming and costly. Consequently, only a small number of fishing trips can be assessed in this way. A recent, more favorable, way to assess discards is to make use of electronic-monitoring (EM) systems \citep{van2020electronic}. In such a system, surveillance cameras are permanently installed on a vessel, allowing the collection of video footage for all trips. However, EM still relies on human observation of the footage, limiting the sample size. A preferred solution to increase the fleet coverage is to install dedicated image-acquisition systems and to develop automatic fish-detection methods \citep[e.g.,]{ovalle2022use, van2021automatic,sokolova2023integrated}. These systems collect high-resolution images of the discarded fish under controlled illumination and typically use deep neural networks (DNN) to detect and classify the catch. This allows assessment of fleet-wide implementation and assess all fishing trips. 

A big challenge for automated discard monitoring using DNNs is the enormous amount of variation present in the camera images. Individual fish, for instance, look differently due to natural variation, age, and seasonal changes. Fish position on the conveyor belt can additionally vary due to overall amount of fish and the  fishers' practices of sorting the catch. Additionally, the catch composition can differ per fishing ground and date.  Moreover, fish often overlap each other creating occlusions that challenge automatic detection and the level of occlusion differs per vessel. Fish detection is particularly difficult for demersal fisheries, as the catch contains a large amount of debris and benthos, which further increases the complexity and variation. To properly deal with these variations and occlusions, it is important to have a large and varied training set covering all situations encountered in practice \citep{ruigrok2023generalization}. While raw camera images can be collected relatively easily, the limiting factor in the collection of training data is the annotation of the images. Annotating the bounding boxes and possibly contour polygons is time consuming and the classification of fish species requires the involvement of experts, making the annotation process expensive. It is therefore important to annotate only those images that are most relevant to the training process, instead of annotating all images or a random subset of the images. 

This challenge is addressed by active learning. Active learning is an iterative process, which lets a deep-learning (DL) model select a set of most relevant images that are then annotated by a human, to be used to retrain the model \citep{monarch2021human}. Key to this process is the method to select the set of most relevant images, which can be based on diversity sampling or uncertainty sampling. Epistemic uncertainty, representing the model uncertainty, can be differentiated from aleatoric uncertainty, representing uncertainty due to poor quality images. Both were shown to be useful to estimate uncertainty of a DNN to detect chickens and estimate their condition \citep{lamping2023uncertainty_plumage}. Uncertainty-based active learning showed to be able to improve model performance with a reduced number of training images in agricultural applications, such as, image-based crop-weed classification \citep{zahidi2021}, semantic segmentation of crop and weed \citep{vanmarrewijk2024}, detection of crop/weed plants \citep{susmelj2024active_learning}, and instance segmentation and disease/defect classification of broccoli \citep{blok2021active}. \cite{blok2021active} estimated epistemic (model) uncertainty with Monte-Carlo dropout during model inference \citep{gal2016dropout}, using this to iteratively select new training data. This resulted in the active learner to have the same performance after sampling 900 images as the random sampling had after 2300 images. However, with the notable exception of \cite{shah2023mi}, these approaches have not been tested in the context of fish detection and monitoring of bycatch in demersal fishery. \cite{shah2023mi} applied multiple instance active learning,  based on adversarial classifiers, which select the most informative fish images from unlabelled images in the context of underwater fish detection. It is, therefore, unclear if uncertainty-based active learning can effectively deal with the large amount of variation and occlusions present in the catch image data to alleviate the annotation burden.

In this paper, we test the hypothesis that the performance of detecting fishes in image of demersal bycatch on a conveyor belt can be improved with fewer training images by making use of active learning. To test the hypothesis, we propose BoxAL\footnote{https://github.com/pieterblok/maskal}, a modification of the Faster R-CNN object detector \citep{ren2015faster}, including epistemic certainty estimation using Monte-Carlo dropout based on \citep{blok2021active, gal2016dropout}. The method selects new training samples for annotation from a large pool of unlabeled images based on three aspects of the model certainty: the occurrence certainty of a fish being present, the semantic certainty of the classified species, and the spatial certainty of the precision of the predicted bounding box. We compared the performance of the active learner with that of a random-sampling strategy. Model training benefits most from new training samples that the model cannot deal with yet. We, therefore, investigated the relation between model certainty and model performance was studied, to see if model certainty indeed results in the selection of those images that the model makes more errors on compared to the remaining samples.   
\section{Materials and Methods}

This section describes the dataset in Section~\ref{subsec:dataset}, the proposed model for active-learning for fish detection, BoxAL, in Section~\ref{subsec:boxal}, the training and sampling procedure in Section~\ref{subsec:training}, and finally the evaluation methods in Section~\ref{subsec:evaluation} 

\subsection{Dataset}
\label{subsec:dataset}
\begin{figure}
    \centering
    \includegraphics[width=0.9\linewidth]{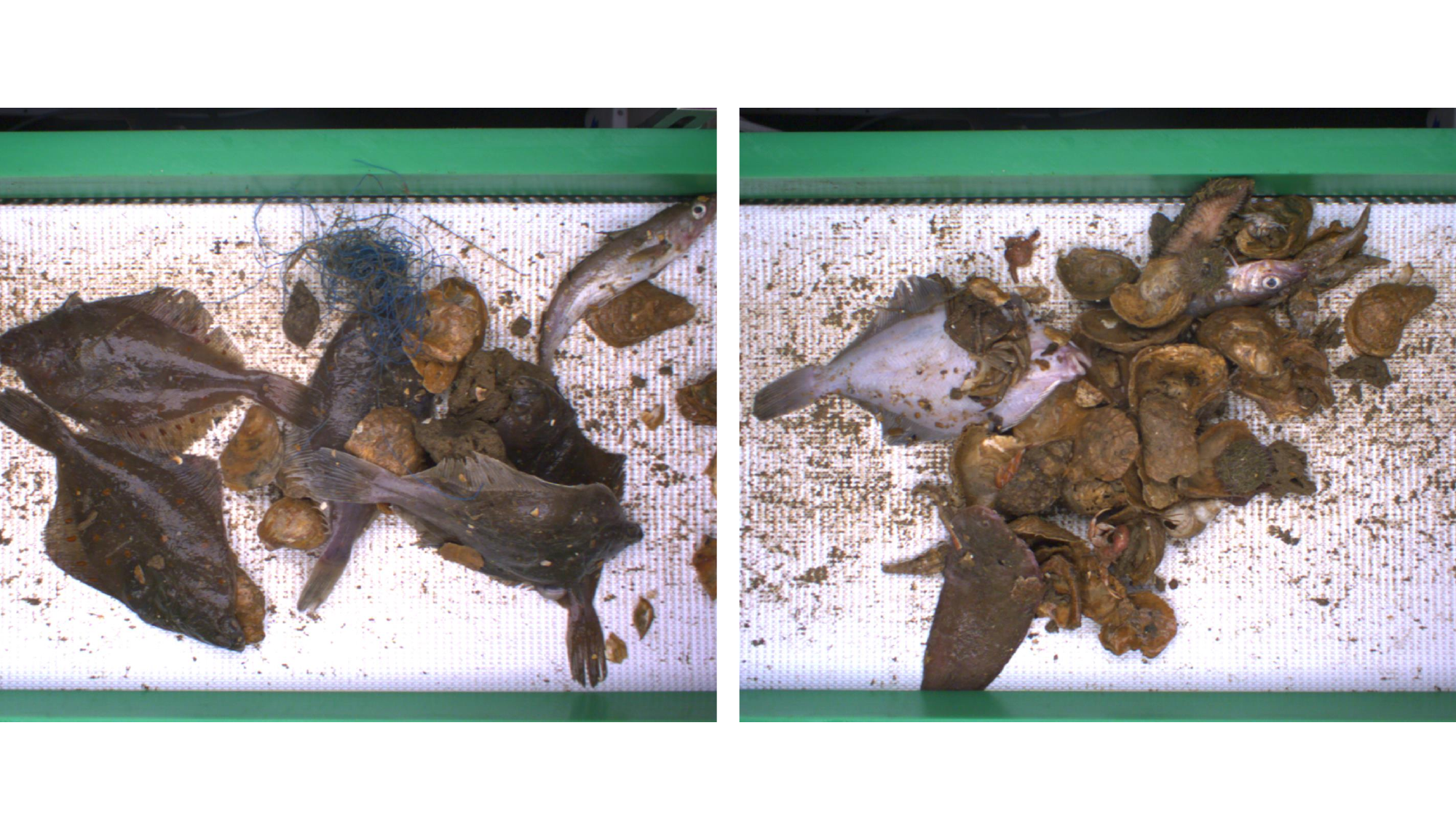}
    \caption{Examples of images from the dataset \citep{https://doi.org/10.4121/16622566.v1}}
    \label{fig:FD_sample}
\end{figure}

We used an open-source dataset with a total of 3005 images of discarded fish by Dutch beam trawlers \citep{https://doi.org/10.4121/16622566.v1}. The dataset contains images of ten demersal fish species commonly discarded by the beam trawlers. Figure \ref{fig:FD_sample} demonstrates a couple of examples from the dataset. The dataset is highly unbalanced and dominated by three flatfish species (\textit{Pleuronectes platessa}, \textit{Solea solea}, and \textit{Limanda limanda}) and whiting (\textit{Merlangius merlangus}). The data contains multiple fishes per image and with different levels of complexity and overlap, as well as debris and benthos.  A summary of the dataset is presented in Table \ref{tab:dataset_summary}, including the split of data in different sets, which will be explained in Section~\ref{subsec:training}.


\begin{table}[ht]
    \centering
    \begin{tabular}{l|l|c|c|c|c|}
        \hline
             \multicolumn{2}{c}{\textbf{Species}}&   \multicolumn{4}{c|}{\textbf{\# Annotations}}\\
          \textbf{English name}&\textbf{Latin name}& \textbf{Initial training} & \textbf{Training} & \textbf{Validation} & \textbf{Test} \\
         \hline
          Starry ray&\textit{Amblyraja radiata} & $32$ & $1092$ & $330$  & $302$ \\
          Grey gurnard&\textit{Eutrigla gurnardus} & $94$ & $263$ & $48$ & $31$\\
          Common dab&\textit{Limanda limanda} & $79$ & $1961$ & $540$ & $322$ \\
          Whiting&\textit{Merlangius merlangus} & $31$ & $3555$ & $988$ & $529$\\
          Lemon sole&\textit{Microstomus kitt} & $23$ & $1004$ & $227$ & $111$\\
          European plaice&\textit{Pleuronectes platessa} & $34$ & $4413$ & $1153$ & $739$\\
          Lesser spotted dogfish&\textit{Scyliorhins canicula} & $70$ & $349$ & $62$ & $44$\\
          Turbot&\textit{Scophthalmus maximus} & $72$ & $194$ & $47$ & $15$\\
          Pouting&\textit{Trisopterus luscus} & $87$ & $153$ & $7$ & $14$\\    
          Common sole&\textit{Solea solea} & $34$ & $3524$ & $901$ & $452$\\
         \hline
    \end{tabular}
    \caption{Dataset summary indicating number of instances per species split into initial train (100 images), train, validation and test subsets.}
    \label{tab:dataset_summary}
\end{table}

\subsection{BoxAL, object detection, and certainty estimation}
\label{subsec:boxal}
\begin{figure}[h!]
    
    \includegraphics[width=\textwidth, center]{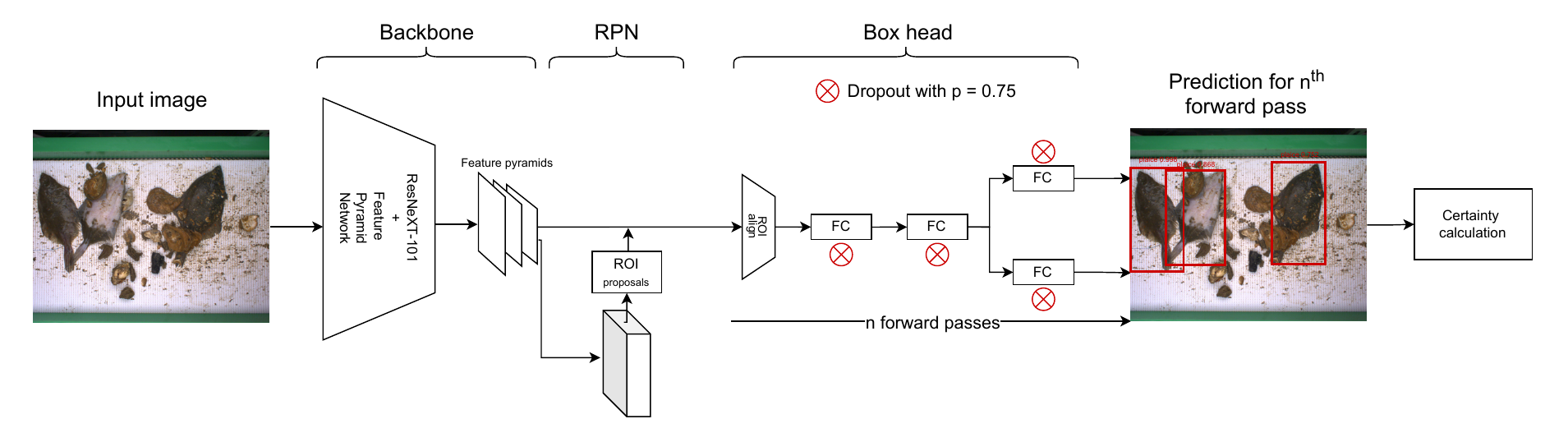}
    \caption{BoxAL architecture with the dropout layer included in the head of the model. During Monte-Carlo inference the drop-out layers are activated in the Box head ($p = 0.75$) and $n$ ($n=15$) forward passes are performed, resulting in $n$ predictions per object in the image. After all forward passes are complete, the predictions per object are then used for semantic, spatial and occurrence certainties calculation.}
    \label{fig:boxal_architecture}
    
\end{figure}

A Faster R-CNN network \citep{ren2015faster} with a ResNeXt-101 \citep{xie2017aggregated} backbone pretrained on ImageNet \citep{deng2009imagenet} has been used as the object-detection model. This two-stage object detector was chosen for its higher detection performance, compared to one-stage detectors \citep{zou2023object}, in the case of large amounts of small and densely distributed objects, which is characteristic for the dataset used in the current study, depicting large numbers of undersized fish of quota-regulated species. The model's backbone calculates a feature pyramid, which is used by the region proposal network (RPN) to propose a large number of regions of interest (ROI) in the image. The ROI-align applies each ROI to the image features, which is subsequently processed by two fully-connected layers, to then be passed on to a fully-connected layer predicting the four bounding box coordinates and to a fully-connected layer predicting the classification scores. The model was modified to estimate epistemic model certainty using Monte-Carlo dropout \citep{gal2016dropout, blok2021active}. This was done by adding random dropout to all fully-connected (FC) layers in the box head, see Figure \ref{fig:boxal_architecture}. The dropout probability has been set to $0.75$. Random dropout was applied during training as a regularization technique. During inference, random dropout was applied to estimate the model certainty. To this end, $n$ forward passes were performed on the box head, each with a different random dropout pattern in the fully-connected layers. Forward passes were performed with fixed thresholds for the non-maximum suppression and confidence, which were set to $0.3$ and $0.5$, respectively. Each forward pass $i$ results in a set of $m_i$ detected object instances, $O_i = \{o_1, \ldots\ o_{m_i}\}$, where each instance $o_j=\{B_j,K_j\}$ consists of four bounding box coordinates prediction, $B_j$ and a vector containing the class probabilities $K_j=\{k_1, \ldots,k_\kappa\}$ for all $\kappa$ fish species. In total, the BoxAL model results in $n$ predictions of objects detected in the image, $\{O_1, \ldots, O_n\}$.  

\label{sec:certainty_estimation}
As in \cite{blok2021active}, from the $n$ forward passes, the instance detections are found that relate to the same object in the environment, resulting in a set of $m$ instance sets, $\mathbb{S}=\{S_1, \ldots, S_m\}$. Each instance set, $S=\{s_i, \ldots,s_r\}$ consists of $r$ instance detections relating to the same object. 
The instance sets are determined starting at the first forward pass, where each object detection in $O_1$ starts an instance set. Next, the detections from the second pass, $O_2$, are matched to the instance sets. If a bounding-box prediction $B_i$ has an intersection-over-union (IoU, Eq. \ref{eq:iou}) of 0.5 or more with at least one bounding-box $B_j$ in the instance set, that is, $\text{IoB}(B_i,B_j)\ge0.5$, the detection is added to the set. If $B_i$ does not match with any instance set, a new instance set is created. This process is repeated for all forward passes, until $O_n$.

\begin{equation}
\label{eq:iou}
    \text{IoU}(B_i, B_j) = \frac{|B_i\cap B_j|}{|B_i\cup B_j|}
\end{equation}
\noindent

After the instance sets are defined, the detections within each set are used to calculate three types of certainty: semantic certainty ($c_{\text{sem}}$), indicating the consistency of the model in class prediction, spatial certainty ($c_{\text{spa}}$), providing the consistency of the model in object localization, and occurrence certainty ($c_{\text{occ}}$), indicating the certainty that the instance set relates to an actual object occurrence. 
The semantic certainty determines how certain the model is about the predicted class/specie and is calculated using entropy, as in \cite{blok2021active}.
First, the entropy for a single instance detection, $s_i$ is calculated based on the associated vector with class probabilities, $K_i$ using: 
\begin{equation}
\label{eq:od_set_entropy}
    H(K_i) = -\sum_{j=1}^{\kappa}{k_j\cdot \log k_j}
\end{equation}
\noindent
The entropy $H(K)$ is normalized between 0 and 1 using the maximum entropy for $\kappa$ number of classes in the object detection set as follows:
\begin{equation}
\label{eq:od_set_max_entropy}
    H_{max}(K_i) = -\kappa \cdot \left(\frac{1}{\kappa}\cdot log\frac{1}{\kappa}\right)
\end{equation}
\noindent
The normalized entropy for the instance detection $s_i$ is calculated as the ratio between $H(K_i)$ and $H_{max}(K_i)$ and the certainty is determined as 1.0 minus the normalized entropy. Then, the semantic certainty $c_{\text{sem}}$ for the instance set $S$ is the average certainty over the instances:
\begin{equation}
\label{eq:c_sem}
    c_{\text{sem}}(S) = \frac{1}{r} \cdot \sum_{i=1}^{r}(1-\left(\frac{H(K_i)}{H_{max}(K_i)}\right) 
\end{equation}

Spatial certainty represents how certain the model is about the bounding-box prediction. The spatial certainty for the instance set, $c_{\text{spa}}$ is defined by comparing all bounding-box predictions in the set to the mean bounding box of the set. The mean box, $\bar{B}(S)$, is determined by the four centroids of the four corner points of the individual boxes in the instance set. The spatial certainty is than calculated as the average IoU between the mean bounding box and each instance bounding box $B_i$ in the set. 
\begin{equation}
\label{eq:c_spl}
    C_{\text{spa}}(S) = \frac{1}{r}\sum_{i=1}^{r}{\text{IoU}(\bar{B}(S), B(s_i))}
\end{equation}

The occurrence certainty $c_{\text{occ}}$ for the instance set was defined as the proportion of instance predictions in the instance set and the total number of forward passes used:
\begin{equation}
\label{eq:c_occ}
    C_{\text{occ}}(S) = \frac{r}{n}
\end{equation}

Finally, the three certainties are combined to get a single certainty value per instance set, using:
\begin{equation}
\label{eq:c_h}
    c_{\text{h}}(S) = c_{\text{sem}}(S) \cdot c_{\text{spa}}(S) \cdot c_{\text{occ}}(S)
\end{equation}
\noindent
These are then combined into a certainty for the whole image by calculating the minimum certainty over all instance sets in $\mathbb{S}=\{S_1,\ldots,S_m\}$:
\begin{equation}
    c_\text{min} = \min_{i=0}^m c_{\text{h}}(S)
\end{equation}

\subsection{Training and sampling procedure}
\label{subsec:training}
Unlike the traditional split of the data into training, validation, and test sets used for supervised learning, for active learning, the training set is adaptive. Initially it consists of a small set of labelled images forming the \textit{initial training set}. The rest of the images in the train set are considered unlabelled and are referred to as the \textit{unlabeled pool}. In this study, the number of potential training images equalled to 3005 images; 100 images, representing all ten species were selected and used as the initial training set (see Table \ref{tab:dataset_summary}). The validation set contained 660 randomly selected images (22\%) and the test set contained 449 randomly selected images (15\%). Two types of augmentations were used during training: random horizontal flip of the image with $0.5$ probability and resizing along the shortest edge of the image while maintaining aspect ratio. Image batch size was set to two. The stochastic gradient descent optimiser was used with a momentum of 0.9 and a weight decay of 1e-4, learning rate was set to 5e-4.

\begin{figure}[h!]
    
    \includegraphics[width=\textwidth, center]{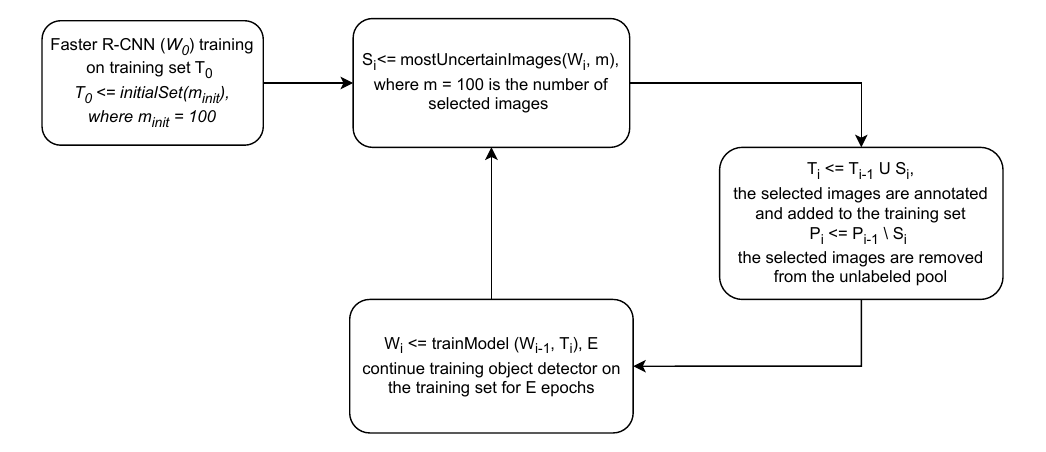}
    \caption{Iterative training procedure of BoxAL. $\mathcal{W}_0 \leftarrow \texttt{initNetwork}()$ (Faster R-CNN with ResNeXt-101 backbone pre-trained on ImageNet). $\textbf{T}_i$ is the set of images in the training set at iteration $i$; $\textbf{P}_i$ is the set of images in the unlabelled pool at iteration $i$; $\textbf{S}_i$ is the set of selected image; $\mathcal{W}_i$ are the weights of the detection network at iteration $i$; $E= E + 5$ epochs, where the initial number of epochs was set to 5.}
    \label{fig:boxal_training}
    
\end{figure}

The active-learning process consists of a number of steps (see also Figure.~\ref{fig:boxal_training}): 1) training on the initial training set, 2) calculation of model certainty for all images in the unlabeled pool and selection of $N$ images that the model is most uncertain about, 3) annotation of the selected images to increase the training set, and 4) retraining of the model on the new training set. Steps 2-4 are iterated until done. Each step is explain in more detail below:
\begin{enumerate}
    \item Faster R-CNN, with ResNeXt-101 backbone pre-trained on ImageNet dataset \citep{deng2009imagenet}, was trained on the initial training set ($m_{init}$), consisting of 100  (Table \ref{tab:dataset_summary}), where the best model weights (Figure \ref{fig:boxal_architecture}) were selected based on the best performance on the validation set. All other hyper-parameters were kept as default, as described in \cite{blok2021active}.
    \item At every iteration all remaining images in the unlabeled pool were processed by BoxAL, with the probability of dropout set to $0.75$ and the number of forward passes to $n=15$ (Figure \ref{fig:boxal_architecture}). The certainty per image was calculated as described in Section~\ref{sec:certainty_estimation}. Next,  $m=100$ images with the lowest certainty value were selected.
    \item The selected 100 images and their corresponding annotations were added to the training set and removed from the unlabeled pool accordingly. 
    \item The neural network was further trained starting with the weights obtained in the previous iteration. The best model weights were selected based on the validation set. 
\end{enumerate}    
As the training set expanded after each active-learning iteration, the number of epochs increased proportionally. The initial number of epochs was set to 5, subsequently, 5 epochs were added after the training set increased with 100 sampled images. 
Steps 2-4 were repeated for 10 active-learning iterations, ending with a training set containing 1,100 images. 

\subsection{Evaluation}
\label{subsec:evaluation}
To test the hypothesis that active learning can result in a better fish detection performance with fewer training images, we compared BoxAL to a baseline approach using \textit{random sampling} in step 2 instead of the uncertainty-based sampling. For the rest, the iterative training of the random-sampling model was identical to training of BoxAL, with the same number of images sampled from the unlabeled pool (100) and 10 iterations.

To measure performance and compare BoxAL to the random-sampling model, after each training iteration the mean average precision (mAP) was calculated using the COCO evaluator\footnote{http://cocodataset.org/detection-eval}. Both training procedures, minimum-certainty and random sampling, have been repeated 5 times to be able to plot the mean performance and the 95\% confidence interval on the mean. The initial training set was set once and fixed for the 5 runs.  As comparison, a baseline Faster R-CNN model was trained on all 3005 training images available with the dropout layers implemented as a regularization technique, but not used during inference.  

In addition, we tested the hypothesis that higher certainty of the model correlates with a higher performance. To this end, at every active-learning iteration, the F1 scores \citep{manning2008introduction} were calculated on the predictions made by the trained BoxAL model on the selected images based on certainty. Per selected image, the associated F1 score and certainty were stored and plotted.  

When new training samples are correctly selected, it is assumed that the performance of the model on the image selected from the unlabeled pool is significantly lower than the performance on the images remaining in training pool. To test this, we performed a two-sided unpaired Student's t-test comparing the F1-scores for the selected set of images to the F1-scores of the remaining images in the pool. The t-test was performed with the SciPy library (v1.12.0) \citep{2020SciPy-NMeth} and the p-values were reported.

\begin{figure}[t!]
    \centering
    \includegraphics[scale = 0.6, center]{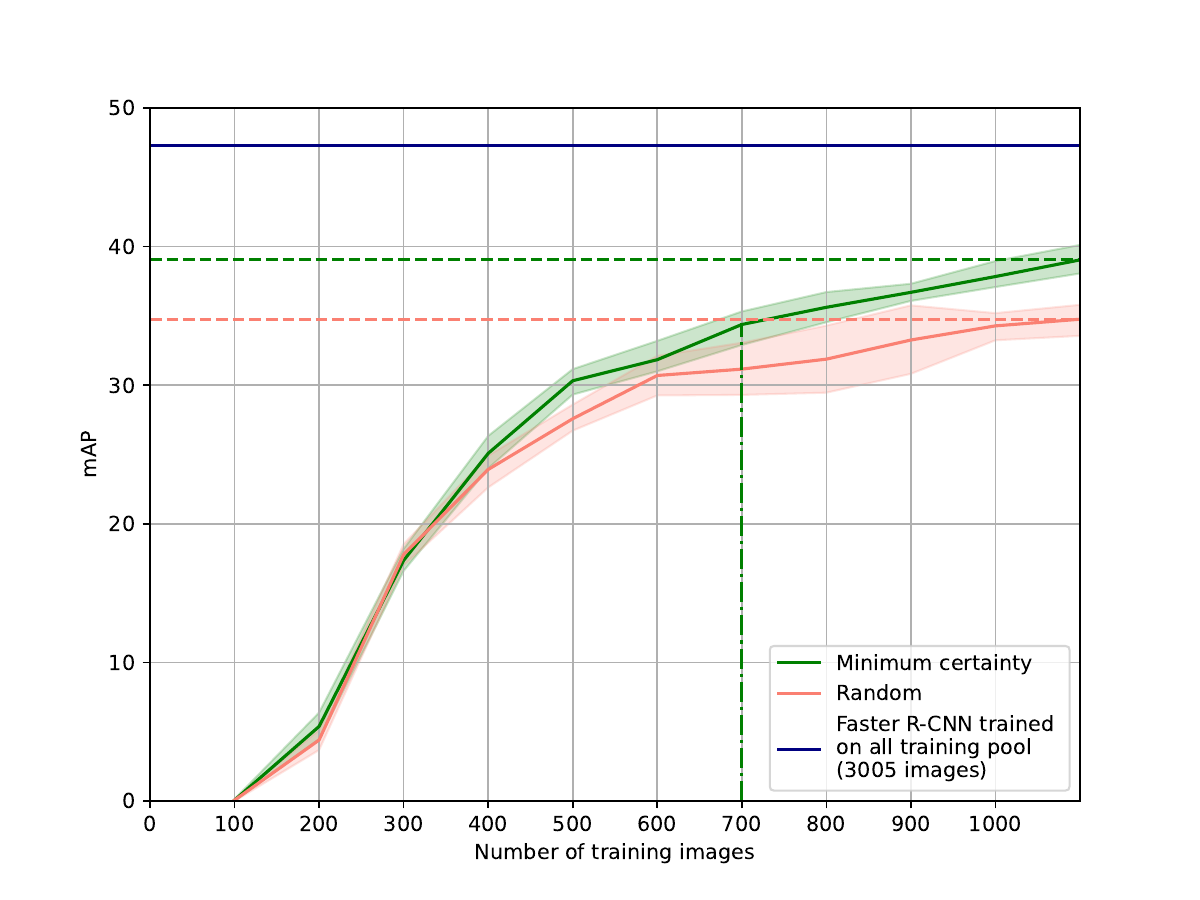}
    \caption{Performance means of the BoxAL active learning with minimum certainty sampling (green solid line) and the random sampling (pink solid line). The coloured areas around the lines represent the 95\% confidence intervals around the means at five repetitions. The blue solid line is the performance of the Faster R-CNN model trained on the entire unlabeled pool (3005 images).}
    \label{fig:min_certainty_vs_random}
\end{figure}

\section{Results}

\subsection{Comparison of BoxAL active learning and random sampling}

This section presents the comparison in the performance between BoxAL trained on images selected on the basis of minimum image certainty and BoxAL trained with randomly sampled images. During the first three training iterations the performance of the two strategies does not differ (Figure \ref{fig:min_certainty_vs_random}). After the fourth training iteration, the difference in performance between the two models starts to diverge. The performance of BoxAL is significantly higher than that of the model trained on the same training size, but with randomly sampled images. Performance of the models from the sixth to the eighth training iterations was not significantly different. However, the performance of BoxAL trained on the images that were sampled using minimum certainty was higher. For the final iterations, the active-learning strategy resulted in significantly higher performance compared to random sampling. The mean mAP after 10 iterations was $39.0\pm{1.6}$ for active-learning training pipeline and $34.8\pm{1.8}$ for the random sampling; this indicates an improvement of  $12.1\%$. It is worth mentioning that the mean mAP of $34.4 \pm{2.0}$ was reached by active learning training pipeline at the sixth training iteration with the training set consisting of 700 images, showing that the same performance can be reached with 400 fewer training samples (Figure \ref{fig:min_certainty_vs_random}). Considering the baseline object-detector performance (mAP) reached with training on all 3005 images was $47.3$, active-learning pipeline reduced the gap in performance between the baseline performance and the performance of random-sampling pipeline by $33.6\%$.

\subsection{Certainty estimation and prediction performance}

\begin{figure}[htb!]
    
    \includegraphics[scale = 0.6, center]{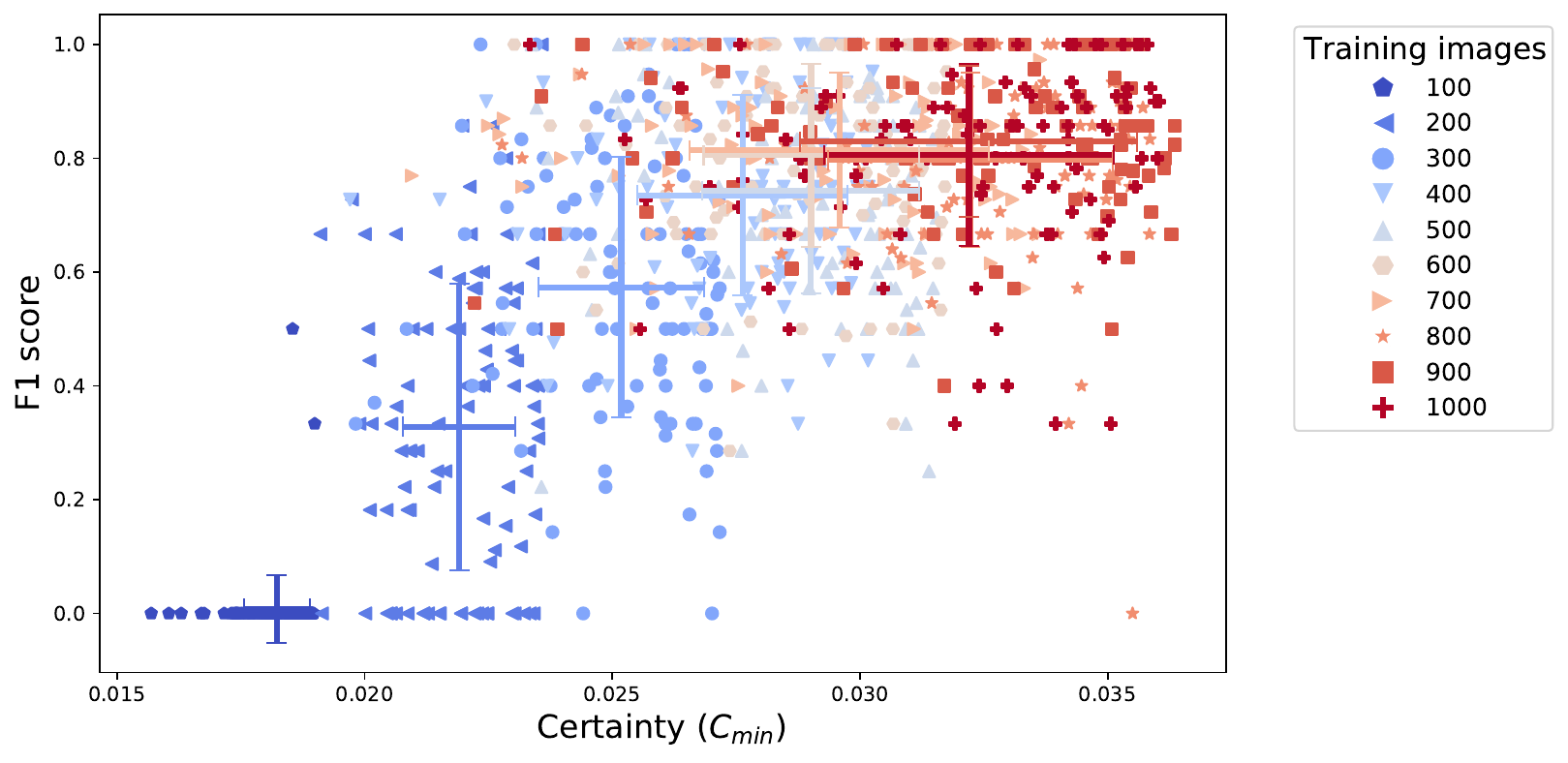}
    \caption{Relationship between F1 scores and certainty values for the training images sampled during ten iterations. Every point in the scatterplot corresponds to calculated F1 score and certainty ($C_{min}$) for a sampled image. Horizontal error bars correspond to certainty standard deviation; vertical error bars correspond to F1 score standard deviation.}
    \label{fig:f1_vs_certainty_sampled}
    
\end{figure}

There is an overall positive correlation between certainty estimation in the consecutive training iterations and F1 score per individual image (Figure \ref{fig:f1_vs_certainty_sampled}). The most rapid increase in F1 score is observed during the second and the third training iterations: the F1 score increases from 0.008 to $0.33$, in the first and the second training iterations. Consequently, the F1 score increases up to $0.6$ in the third training iteration. After the fourth iteration the F1 scores start to plateau and at the final iterations the mean F1 score increases from $0.74$ in the fourth iteration to $0.8$ in the last iteration. Positive correlation is observed over all iterations and the method becomes more certain over time.

\subsection{Comparison of F1 scores of the sampled images and the images remaining in the unlabeled pool}

\begin{figure}[htb!]
    
    \includegraphics[scale = 0.6, left]{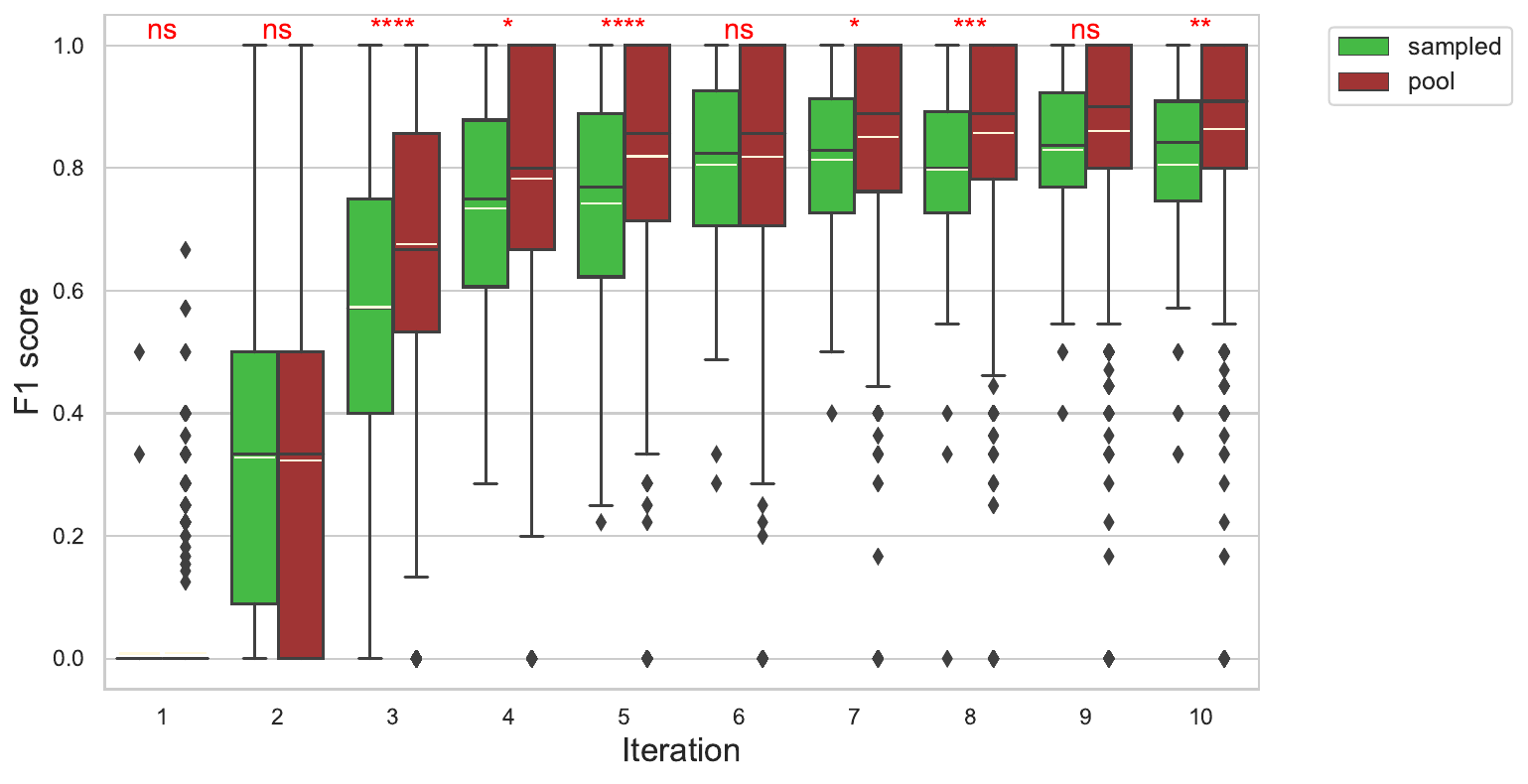}
    \caption{Comparison of F1 scores of the sampled and remaining in unlabeled pool images. The yellow marker corresponds to the mean F1 score value. Statistical significance is indicated as follows: ns: $p > 0.05$; $*$: $p \leq 0.05$; $**$: $p \leq 0.01$; $***: p \leq 0.001$; $****$: $p \leq 0.0001$.}
    \label{fig:f1_pull_not_pull}
    
\end{figure}

We hypothesise, that active-learning pipeline should select the images that the method handles poorly. This is indicated with lower F1 scores for the selected images compared to the ones remaining in the unlabeled pool. Figure \ref{fig:f1_pull_not_pull} presents the comparison of F1 scores of the images that were sampled by BoxAL and the images that remained in the unlabeled pool in each of the training iterations. During the first two training iterations the difference between the mean F1 score of the sampled images and the images remaining in the unlabeled pool was not significant. 
In the subsequent training iterations, from the third to the fifth and in the seventh, eighth and tenth iteration the F1 score of sampled images was significantly lower than the F1 score of the images remaining in the unlabeled pool. The resulting F1 score in the final training iteration reached the  F1 score of $0.81$ for the sampled images and $0.86$ for the images remaining in the unlabeled pool. Notably, 26 images that remained in the unlabeled pool in the tenth iteration had an F1 score of 0, while there were no images with zero F1 score among the sampled images in the final training iteration.  
In ideal scenario, our method would always select the images with lower F1 scores, which would be indicated by significantly lower F1 scores of the sampled images compared to the remaining in the pool among all iterations. However, Figure \ref{fig:f1_pull_not_pull} shows that there is a substantial overlap between the F1 scores of sampled and remaining in pool images. Additionally, the images with high F1 scores were sampled from the second training iteration, while the images with low F1 scores were not selected by the method.  

\section{Discussion}

\subsection{BoxAL comparison with related studies in the field of natural science and medicine }

The current study demonstrated that active learning helps to select a more informative training set that allows the object-detector to generalize better using 400 less training images for the case of fish discards documentation. Additionally, BoxAL perfomed $12.1\%$ better than random-sampling pipeline and minimized the gap in performance between the baseline performance and the performance of random-sampling pipeline by $33.6\%$.
Our finding is consistent with related studies, where authors applied active-learning in the fields of agriculture, medical imaging, and underwater monitoring. The study by \cite{blok2021active} demonstrated that active learning based on MaskAL reaches significantly higher performance during all training iterations than random sampling. The dataset used in that study contained a 4.6 times larger dataset in the training pool than in the current study. The images showed broccoli heads, where the heads were classified with four different diseases and defects, with on average two objects per image filmed from a top view. The objects' appearance was, therefore, more consistent, since only one specie was present, and broccoli are planted consistently in a row in the field, and broccoli heads do not change their position in respect to camera, which is in contrast to the appearance of different fish species in the dataset used for the current study.  Yet, in our study with the higher variation and occlusion, active learning resulted in a higher model performance. This happened after the fifth training iteration, which can be explained by the model learning enough about the representation of different fish species to be able to select uncertain cases.
Active-learning based on uncertainty estimation for the image classification task has been described in the medical field \citep{leibig2017leveraging}. The authors applied the principle of Bayesian approximation described in \cite{gal2016dropout} to solve a binary classification task of fundus images into 'healthy' and 'diseased' categories. Several custom CNN architectures were compared. Similarly to our study,  authors showed that model uncertainty was significantly higher for incorrect predictions. Since the authors worked with a classification network, their method of uncertainty estimation can be compared with the semantic certainty ($C_{sem}$) in our study. 
\cite{bochinski2019deep} also proposed using active learning for classification task in the field of biodiversity monitoring, specifically zooplankton. The aim of the study was to apply active learning to leverage the performance of deep classification networks with fewer training images. The authors used a classification CNN and applied a cost-effective active learning approach. Their approach utilizes three types of uncertainty estimation: least confidence, margin sampling and entropy. The approach proposed by \cite{bochinski2019deep} focuses on sampling the most informative images, approximately indicated by the highest model uncertainty, and also uses the high-confidence predictions of the majority classes. This results in a dataset where the most uncertain images are manually labeled by human supervisors, and the images containing objects classified with high confidence are added to the training set with pseudo-labels. High confidence predictions can also serve as preudo-labels during training in the current study, this will further minimize annotation effort.

\subsection{Active learning for object detection}
Active-learning paradigm for object detection has been introduced from early 2000s. In those studies object detection was performed based on the pre-selected features, that  were found to be specific for the object of interest.
For instance, a study by \cite{abramson2006active} describes an active learning approach for pedestrians detection. The authors pre-selected a illumination-independent visual feature in the images that was highly correlated with pedestrians present in the image. Authors used AdaBoost to predict a score for each candidate sample, then the samples close to decision boundary would be selected as the most informative ones.  Authors reported the improvement of the object detector after every training iteration with the selected samples based on active-learning.

Concerning deep-learning CNNs, the majority of active-learning approaches focused on implementing model certainty estimation in classification networks. Studies by \cite{wang2014new} describe active-learning pipeline for image classification tasks based on least confidence, margin and entropy. \cite{wang2016cost} proposed an extension of the active-learning pipeline for image classification by utilizing high-confidence predictions as pseudolabels. \cite{stark2015captcha} margin sampling for captcha classification. All of these studies demonstrated an advantage of using active-learning in comparison with random sampling.
Implementing certainty estimation on more complex models, such as object-detectors is not a trivial task, as it is often associated with high computational load and management of the multiple predictions per image. 
\cite{brust2018active}  proposed active learning for YOLOs object detector \citep{redmon2016you}.  Despite the proposed active-learning pipeline is applied to an object detector, the selection of the most informative images was based on margin sampling of class confidence scores. Our study expanded the criteria for active-learning image sampling  with  spatial and occurrence certainties in addition to the semantic certainty. 

\subsection{Considerations of BoxAL deployment and future work}
In the current study, the full image was used as input for the model. When the minimum over all detections certainty is calculated it does not eliminate the fact that the other class that has already been learned by the detector is present in the image, while still selected as training data as these objects are in the same image as the detected objects with minimum certainty.  Thus, the informativeness of the added images may be not so high and therefore less beneficial for the model performance improvement. In the future work, it might be beneficial to add only the fish instance with low certainty instead of the full image.

The initial training was created by 100 images that contained all classes present in the full dataset. In the case an unseen class is present in the unlabeled pool images, it is likely not detected in any of the forward passes and does therefore not contribute to the certainty estimation. Thus, unseen species, that are potentially of high informativeness for the model, will not be selected. An additional criterion for selection that favours image containing novel content needs to be studies in the future work. One of the approaches that can serve as such criterion is open-set recognition \cite{vaze2022opensetrecognitiongoodclosedset}.
During the first three training iterations, active learning and random sampling do not show a difference in performance. This can be explained by the variation, which implies that any image is informative despite being randomly sampled. Nonetheless, after these initial training BoxAL learned to deal with part of the data and consequently, favoured the training samples selected based on minimum certainty.
To reach the same detection performance as the random sampling, the BoxAL active learning requires 400 less annotated images on a total of 1100. Considering the time needed for annotating an image this can result in saving up to 400 hours of annotation. 
\section{Conclusion}

The proposed active learning approach based on BoxAL allows reaching the same object detection performance using $700$ training images, compared to $1100$ training images used by random sampling pipeline. Further analysis showed that the increase in overall image certainty has a positive correlation with object-detection performance during the training iterations. That supports the hypothesis that the images sampled with the minimum certainty criteria are the images with which the model struggles the most with making them more informative and, therefore, contributing to the increased performance of the fish detector. Additionally, this hypothesis is supported by observing  lower F1 scores of the sampled images compared to the F1 scores of the images remaining in the unlabeled pool. 
The proposed active-learning approach allows for reduction in labor intensity, time and costs needed to annotate the data, while increasing the performance of fish detection in a challenging dataset containing high level of variation and occlusion. This advances the applications of the fish detection to automatically detect the catch composition onboard fishing vessels, supporting the transformation of the landing obligation to documentation obligation. 
\begin{notes}[Acknowledgements]
This study was done under the Fully Documented Fisheries project funded by the European Maritime and Fisheries Fund (contract number 16302).

\end{notes}

\section*{Data availability statement}
The data underlying this article is open-source and available at https://doi.org/10.4121/16622566.v1 \citep{https://doi.org/10.4121/16622566.v1}.  Our software is available on \url{https://github.com/pieterblok/boxal}.

\bibliographystyle{apalike}
\bibliography{bib/bibliography}

\begin{thebibliography}{}

\bibitem[Abramson and Freund, 2006]{abramson2006active}
Abramson, Y. and Freund, Y. (2006).
\newblock Active learning for visual object detection.

\bibitem[Blok et~al., 2022]{blok2021active}
Blok, P.~M., Kootstra, G., Elghor, H.~E., Diallo, B., {van Evert}, F.~K., and {van Henten}, E.~J. (2022).
\newblock Active learning with maskal reduces annotation effort for training mask r-cnn on a broccoli dataset with visually similar classes.
\newblock {\em Computers and Electronics in Agriculture}, 197:106917.

\bibitem[Bochinski et~al., 2019]{bochinski2019deep}
Bochinski, E., Bacha, G., Eiselein, V., Walles, T.~J., Nejstgaard, J.~C., and Sikora, T. (2019).
\newblock Deep active learning for in situ plankton classification.
\newblock In {\em Pattern Recognition and Information Forensics: ICPR 2018 International Workshops, CVAUI, IWCF, and MIPPSNA, Beijing, China, August 20-24, 2018, Revised Selected Papers 24}, pages 5--15. Springer.

\bibitem[Brust et~al., 2018]{brust2018active}
Brust, C.-A., K{\"a}ding, C., and Denzler, J. (2018).
\newblock Active learning for deep object detection.
\newblock {\em arXiv preprint arXiv:1809.09875}.

\bibitem[Deng et~al., 2009]{deng2009imagenet}
Deng, J., Dong, W., Socher, R., Li, L.-J., Li, K., and Fei-Fei, L. (2009).
\newblock Imagenet: A large-scale hierarchical image database.
\newblock In {\em 2009 IEEE conference on computer vision and pattern recognition}, pages 248--255. Ieee.

\bibitem[Essen et~al., 2021]{https://doi.org/10.4121/16622566.v1}
Essen, van, R., Vroegop, A., Mencarelli, A.~A., van Helmond, A., Nyugen, L., Batsleer, J., Poos, J. J.~J., and Kootstra, G. (2021).
\newblock Data underlying the publication: Automatic discard registration in cluttered environments using deep learning and object tracking: class imbalance, occlusion, and a comparison to human review.

\bibitem[EU, 2015]{EU2015}
EU (2015).
\newblock Regulation (eu) 2015/812 of the european parliament and of the council of 20 may 2015 amending council regulations (ec) no 850/98, (ec) no 2187/2005, (ec) no 1967/2006, (ec) no 1098/2007, (ec) no 254/2002, (ec) no 2347/2002 and (ec) no 1224/2009, and regulations (eu) no 1379/2013 and (eu) no 1380/2013 of the european parliament and of the council, as regards the landing obligation, and repealing council regulation (ec) no 1434/98.
\newblock {\em Official Journal of the European Communities}.

\bibitem[Gal and Ghahramani, 2016]{gal2016dropout}
Gal, Y. and Ghahramani, Z. (2016).
\newblock Dropout as a bayesian approximation: Representing model uncertainty in deep learning.
\newblock In {\em international conference on machine learning}, pages 1050--1059. PMLR.

\bibitem[Lamping et~al., 2023]{lamping2023uncertainty_plumage}
Lamping, C., Kootstra, G., and Derks, M. (2023).
\newblock Uncertainty estimation for deep neural networks to improve the assessment of plumage conditions of chickens.
\newblock {\em Smart Agricultural Technology}, 5:100308.

\bibitem[Leibig et~al., 2017]{leibig2017leveraging}
Leibig, C., Allken, V., Ayhan, M.~S., Berens, P., and Wahl, S. (2017).
\newblock Leveraging uncertainty information from deep neural networks for disease detection.
\newblock {\em Scientific reports}, 7(1):1--14.

\bibitem[Manning, 2008]{manning2008introduction}
Manning, C.~D. (2008).
\newblock {\em Introduction to information retrieval}.
\newblock Syngress Publishing,.

\bibitem[Monarch, 2021]{monarch2021human}
Monarch, R.~M. (2021).
\newblock {\em Human-in-the-Loop Machine Learning: Active learning and annotation for human-centered AI}.
\newblock Simon and Schuster.

\bibitem[Ovalle et~al., 2022]{ovalle2022use}
Ovalle, J.~C., Vilas, C., and Antelo, L.~T. (2022).
\newblock On the use of deep learning for fish species recognition and quantification on board fishing vessels.
\newblock {\em Marine Policy}, 139:105015.

\bibitem[Poos et~al., 2013]{poos2013estimating}
Poos, J., Aarts, G., Vandemaele, S., Willems, W., Bolle, L., and Van~Helmond, A. (2013).
\newblock Estimating spatial and temporal variability of juvenile north sea plaice from opportunistic data.
\newblock {\em Journal of sea research}, 75:118--128.

\bibitem[Redmon et~al., 2016]{redmon2016you}
Redmon, J., Divvala, S., Girshick, R., and Farhadi, A. (2016).
\newblock You only look once: Unified, real-time object detection.
\newblock In {\em Proceedings of the IEEE conference on computer vision and pattern recognition}, pages 779--788.

\bibitem[Ren et~al., 2015]{ren2015faster}
Ren, S., He, K., Girshick, R., and Sun, J. (2015).
\newblock Faster r-cnn: Towards real-time object detection with region proposal networks.
\newblock {\em Advances in neural information processing systems}, 28.

\bibitem[Ruigrok et~al., 2023]{ruigrok2023generalization}
Ruigrok, T., van Henten, E., and Kootstra, G. (2023).
\newblock Improved generalization of a plant-detection model for precision weed control.
\newblock {\em Computers and Electronics in Agriculture}, 204:107554.

\bibitem[Shah et~al., 2023]{shah2023mi}
Shah, C., Alaba, S.~Y., Nabi, M., Caillouet, R., Prior, J., Campbell, M., Wallace, F., Ball, J.~E., and Moorhead, R. (2023).
\newblock Mi-afr: multiple instance active learning-based approach for fish species recognition in underwater environments.
\newblock In {\em Ocean Sensing and Monitoring XV}, volume 12543, pages 227--238. SPIE.

\bibitem[Sokolova et~al., 2023]{sokolova2023integrated}
Sokolova, M., Cordova, M., Nap, H., van Helmond, A., Mans, M., Vroegop, A., Mencarelli, A., and Kootstra, G. (2023).
\newblock An integrated end-to-end deep neural network for automated detection of discarded fish species and their weight estimation.
\newblock {\em ICES Journal of Marine Science}, 80(7):1911--1922.

\bibitem[Stark et~al., 2015]{stark2015captcha}
Stark, F., Haz{\i}rbas, C., Triebel, R., and Cremers, D. (2015).
\newblock Captcha recognition with active deep learning.
\newblock In {\em Workshop new challenges in neural computation}, volume 2015, page~94. Citeseer.

\bibitem[Susmelj, 2024]{susmelj2024active_learning}
Susmelj, I. (2024).
\newblock Active learning strategies compared for yolov8 on lincolnbeet.

\bibitem[van Essen et~al., 2021]{van2021automatic}
van Essen, R., Mencarelli, A., van Helmond, A., Nguyen, L., Batsleer, J., Poos, J.-J., and Kootstra, G. (2021).
\newblock Automatic discard registration in cluttered environments using deep learning and object tracking: class imbalance, occlusion, and a comparison to human review.
\newblock {\em ICES Journal of Marine Science}, 78(10):3834--3846.

\bibitem[van Helmond et~al., 2020]{van2020electronic}
van Helmond, A.~T., Mortensen, L.~O., Plet-Hansen, K.~S., Ulrich, C., Needle, C.~L., Oesterwind, D., Kindt-Larsen, L., Catchpole, T., Mangi, S., Zimmermann, C., et~al. (2020).
\newblock Electronic monitoring in fisheries: lessons from global experiences and future opportunities.
\newblock {\em Fish and Fisheries}, 21(1):162--189.

\bibitem[van Marrewijk et~al., 2024]{vanmarrewijk2024}
van Marrewijk, B.~M., Dandjinou, C., Rustia, D. J.~A., Gonzalez, N.~F., Diallo, B., Dias, J., Melki, P., and Blok, P.~M. (2024).
\newblock Active learning for efficient annotation in precision agriculture: a use-case on crop-weed semantic segmentation.

\bibitem[Vaze et~al., 2022]{vaze2022opensetrecognitiongoodclosedset}
Vaze, S., Han, K., Vedaldi, A., and Zisserman, A. (2022).
\newblock Open-set recognition: a good closed-set classifier is all you need?

\bibitem[Virtanen et~al., 2020]{2020SciPy-NMeth}
Virtanen, P., Gommers, R., Oliphant, T.~E., Haberland, M., Reddy, T., Cournapeau, D., Burovski, E., Peterson, P., Weckesser, W., Bright, J., {van der Walt}, S.~J., Brett, M., Wilson, J., Millman, K.~J., Mayorov, N., Nelson, A. R.~J., Jones, E., Kern, R., Larson, E., Carey, C.~J., Polat, {\.I}., Feng, Y., Moore, E.~W., {VanderPlas}, J., Laxalde, D., Perktold, J., Cimrman, R., Henriksen, I., Quintero, E.~A., Harris, C.~R., Archibald, A.~M., Ribeiro, A.~H., Pedregosa, F., {van Mulbregt}, P., and {SciPy 1.0 Contributors} (2020).
\newblock {{SciPy} 1.0: Fundamental Algorithms for Scientific Computing in Python}.
\newblock {\em Nature Methods}, 17:261--272.

\bibitem[Wang and Shang, 2014]{wang2014new}
Wang, D. and Shang, Y. (2014).
\newblock A new active labeling method for deep learning.
\newblock In {\em 2014 International joint conference on neural networks (IJCNN)}, pages 112--119. IEEE.

\bibitem[Wang et~al., 2016]{wang2016cost}
Wang, K., Zhang, D., Li, Y., Zhang, R., and Lin, L. (2016).
\newblock Cost-effective active learning for deep image classification.
\newblock {\em IEEE Transactions on Circuits and Systems for Video Technology}, 27(12):2591--2600.

\bibitem[Xie et~al., 2017]{xie2017aggregated}
Xie, S., Girshick, R., Doll{\'a}r, P., Tu, Z., and He, K. (2017).
\newblock Aggregated residual transformations for deep neural networks.
\newblock In {\em Proceedings of the IEEE conference on computer vision and pattern recognition}, pages 1492--1500.

\bibitem[Zahidi and Cielniak, 2021]{zahidi2021}
Zahidi, U.~A. and Cielniak, G. (2021).
\newblock Active learning for crop-weed discrimination by image classification from convolutional neural network’s feature pyramid levels.
\newblock In {\em Computer Vision Systems: 13th International Conference, ICVS 2021, Virtual Event, September 22-24, 2021, Proceedings}, page 245–257, Berlin, Heidelberg. Springer-Verlag.

\bibitem[Zou et~al., 2023]{zou2023object}
Zou, Z., Chen, K., Shi, Z., Guo, Y., and Ye, J. (2023).
\newblock Object detection in 20 years: A survey.
\newblock {\em Proceedings of the IEEE}.

\end{thebibliography}

\end{document}